\pdfoutput=1

\documentclass[11pt]{article}

\usepackage[]{EMNLP2023}
\usepackage{times}
\usepackage{latexsym}
\usepackage{booktabs}
\usepackage{longtable}
\usepackage{graphicx}
\usepackage{xcolor}
\usepackage{textcomp}

\usepackage[T1]{fontenc}

\usepackage[utf8]{inputenc}

\usepackage{microtype}

\usepackage{inconsolata}

\title{Speakerly\textsuperscript{\texttrademark}: A Voice-based Writing Assistant for Text Composition} 

\author{\textbf{Dhruv Kumar$^{\ast}$, Vipul Raheja$^{\ast}$, Alice Kaiser-Schatzlein, Robyn Perry,} \\
\textbf{Apurva Joshi, Justin Hugues-Nuger, Samuel Lou, Navid Chowdhury} \\
Grammarly \\
\texttt{firstname.lastname@grammarly.com}}

\begin{document}

\newcommand{\speakerly}{Speakerly\textsuperscript{\texttrademark}}
\newcommand{\speakerlysp}{Speakerly\textsuperscript{\texttrademark}\space}

\maketitle
\begingroup\def\thefootnote{*}\footnotetext{Equal contribution by both authors.}\endgroup
\begin{abstract}
We present \speakerly, a new real-time voice-based writing assistance system that helps users with text composition across various use cases such as emails, instant messages, and notes. The user can interact with the system through instructions or dictation, and the system generates a well-formatted and coherent document. We describe the system architecture and detail how we address the various challenges while building and deploying such a system at scale. More specifically, our system uses a combination of small, task-specific models as well as pre-trained language models for fast and effective text composition while supporting a variety of input modes for better usability.
\end{abstract}

\section{Introduction}

Writing is a multi-step process that involves planning (ideation), translation (composition), and reviewing (revision) \cite{f508427a-e4c0-3d6a-8abf-03a5d21ec6c4}. In the ideation phase, the writer gathers information and organizes their thoughts. The composition step involves articulating the ideas effectively through the use of the right words and arranging them cogently in a draft. During revision, the focus is on grammatical correctness, logical flow of ideas, coherent document structure, and style.

Most current writing assistants have been limited in their ability to provide seamless writing assistance across all the stages, take into account the user context, and be robust to work on diverse real-world use cases at scale \cite{gero-etal-2022-design}.

In this work, we introduce \speakerly, a voice-based end-to-end writing assistance system that works across the different stages of writing, helping users become more efficient with their communication. 
The user uses the voice interface to articulate their thoughts in natural speech.
Our system then creates a polished and ready-to-send first draft while addressing all the intermediate issues, such as structure, formatting, appropriate word usage, and document coherence.

\begin{figure}[t]
\includegraphics[width=\linewidth,trim={0 4.2cm 0 4.2cm},clip]{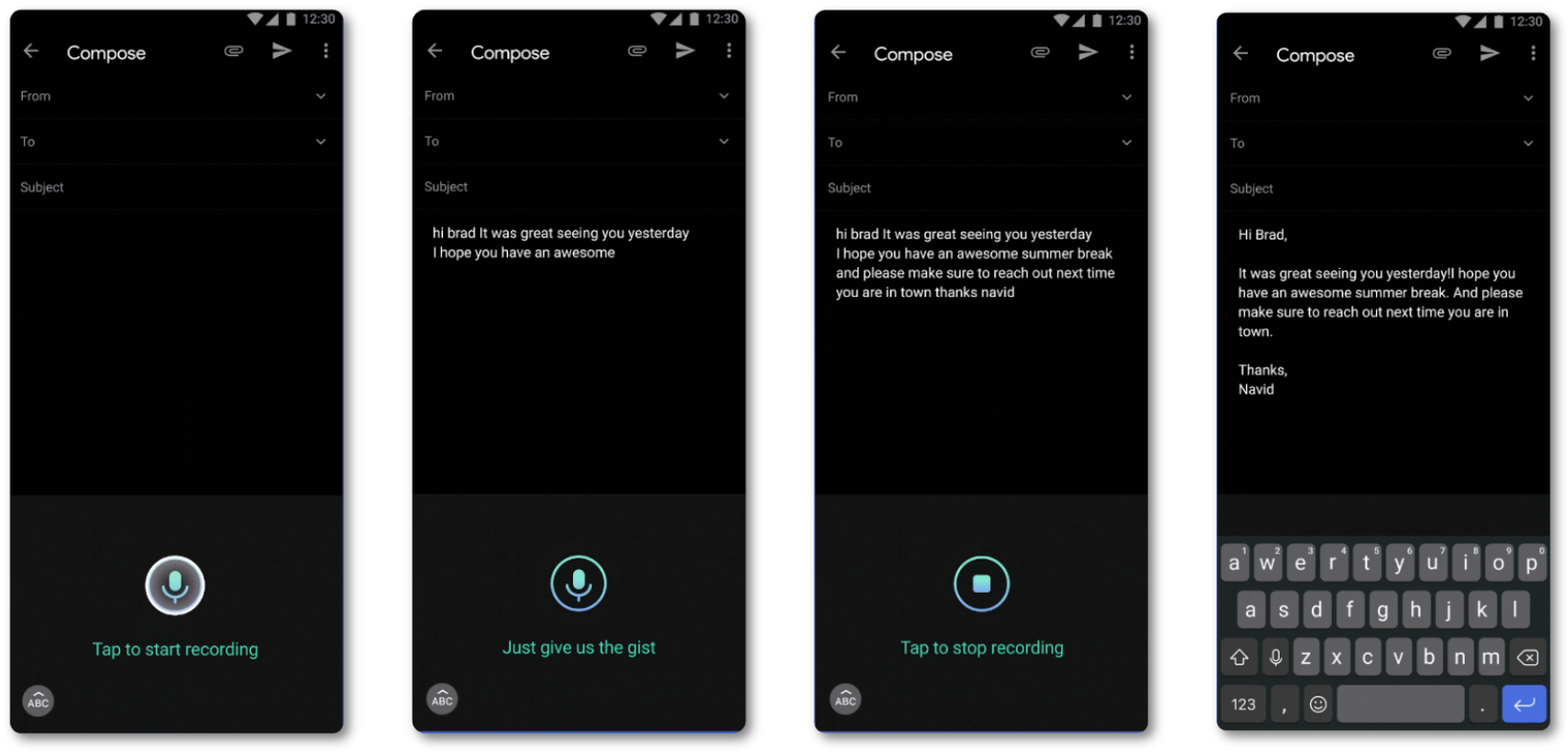}
\centering
\caption{An illustrative example of \speakerlysp for email composition on mobile. A user presses the microphone button at the bottom of their email application and starts speaking naturally (no templatization or structural tailoring of the speech input is needed). Once they stop speaking, \speakerlysp converts the speech into structured, well-formatted, and polished compositions.}
\label{fig:product}
\end{figure}

We use voice as it is a natural and efficient input modality, allowing users to compose their thoughts quickly and even use the system in eyes-free scenarios while performing tasks such as walking and driving \cite{doi:10.1073/pnas.92.22.10031, doi:10.1073/pnas.92.22.9921, 10.1145/3161187}. Moreover, with the increased ubiquity of voice-based assistants, such as Alexa and Siri, voice-based interactions have become more common and intuitive for users \cite{10.1145/3173574.3174214}. 

However, using voice has some challenges. 
First, during the ideation stage, the user typically only has a rough idea of what they want to write. 
Thus, if the system is unable to handle a lack of structure and slight incoherence in the input, users will end up spending a significant amount of time on fixing the output. Second, different writers can have varied needs requiring the system to handle the demands and constraints of different use cases. For example, short vs. long inputs, instructional vs. dictation inputs, open vs closed-ended inputs, and specific structures and formatting for emails, instant messages, and notes (Table \ref{tab:dataset}). Finally, the system should be reasonably fast so that it can provide a delightful user experience.

Speakerly\textsuperscript{\texttrademark} is composed of multiple stages (Section \ref{sec:system_description}) that progressively refine the relatively noisy and unstructured speech from the user and address the aforementioned challenges and requirements. 
In the remainder of this paper, we describe the technical system architecture and our approaches to address challenges related to modeling, evaluation, inference, and sensitivity.

\begin{figure*}[t]
\includegraphics[width=\linewidth,trim={1cm 0.2cm 1cm 0cm},clip]{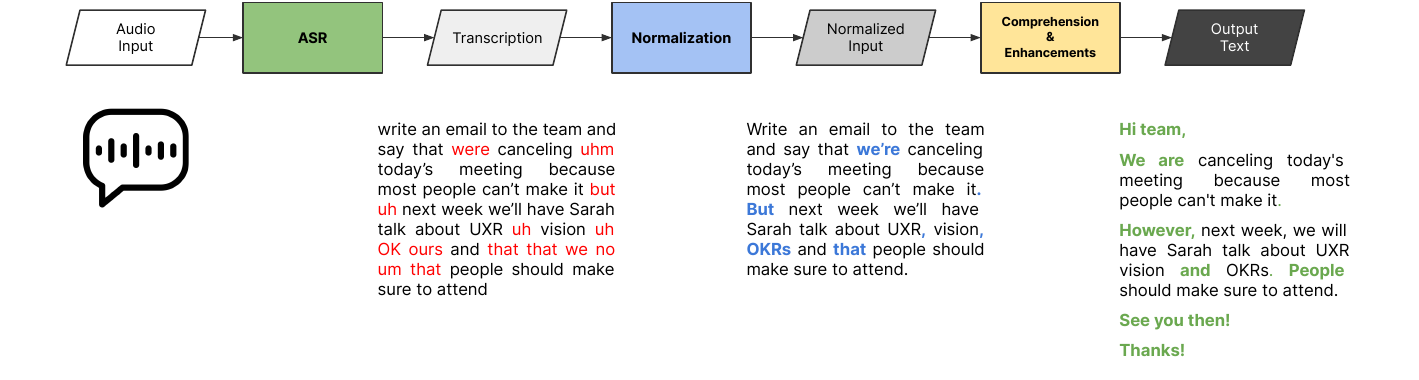}
\centering
\caption{Overview of the system architecture. The ASR system first transcribes the input. Then, the Normalization stage fixes the issues in the transcribed input (shown in red and blue). Finally, the comprehension stage generates a well-formatted and coherent output text with further enhancements.}
\label{fig:architecture}
\end{figure*}

\section{Related Work}

Most research in the past has been limited to either a single use case for composition or one particular stage of the writing process. For example, previous works have focused on email writing \cite{hui2018introassist}, science writing \cite{gero-etal-2022-sparks}, story writing \cite{clark2018creative, coenen2021wordcraft}, slogan and metaphor writing \cite{gero2019metaphoria}, poetry writing \cite{chakrabarty-etal-2022-help}, and support comments \cite{peng2020exploring}, to name a few. Our system, in contrast, can handle various use cases ranging from short instant messages to long notes to open-ended instructions to closed-ended and information-dense dictations.

On the other hand, some writing assistance-focused works disproportionately emphasize specific stages of writing, such as editing and revision \cite{mallinson-etal-2022-edit5, du-etal-2022-read, kim-etal-2022-improving, schick2023peer, raheja2023coedit} rather than end-to-end writing assistance. Again, in contrast, our system is much more extensive as it takes in noisy and unstructured speech input and iteratively refines it to produce a final well-formatted output rather than focusing on a single-shot, structured text-to-text transformation.

Voice-based input has been known to optimize people's interaction and has been studied in the past \cite{williams1998guidelines} and is well-integrated in virtual assistants (such as Siri and Alexa). It has been used for various tasks such as voice notes \cite{stifelman1993voicenotes}, 
data capturing \cite{luo2021foodscrap}, information querying \cite{schalkwyk2010your} and data exploration \cite{srinivasan2020inchorus}. Such systems can have speech recognition errors that are difficult to recover from and restrict the user's natural speaking behavior \cite{luo2020tandemtrack}. 
To tackle these problems, recent works have looked at voice-based text editing \cite{ghosh2020voice,fan2021just}.

\section{System Description}
\label{sec:system_description}

Our system takes natural speech from the user as input and generates a coherent and well-formatted text output. 
As shown in Fig. \ref{fig:architecture}, the input progressively gets refined and enhanced as it traverses the pipeline, consisting of multiple task-specific models. Each stage can have its own errors. Hence, models across the pipeline are designed with complementary, sometimes overlapping capabilities, which allows them to recover from errors collectively and improve robustness to variation and noise in the input. 

The pipeline has three main components: Automatic Speech Recognition (ASR), Normalization, and Comprehension. The ASR module takes raw speech and converts it to text. Then, the normalization module cleans up speech disfluencies, adds punctuation, and applies grammatical error corrections (GEC). 
Finally, the comprehension module cleans the text of remaining issues, such as incoherent document structure, word choice, formatting, formality, and style, and composes the final output text, handling instruction or dictation, or any other mode of input for a variety of use cases. We now explain these three components in more detail:



\subsection{Automatic Speech Recognition (ASR)}
\label{sec:asr_description}
The entry point to the system is an ASR component. This stage is responsible for the transcription of the user's spoken input and also handles basic speech recognition errors, such as filler words and background noise \footnote{We tested the system by using muffled voice and playing street sounds in the background, while speaking to the system. We found that while ASR can deal with most ambient and low noises, any loud sound, can prevent it from picking up some words or inserting incorrect words. We deal with such cases in the Normalization step (Section \ref{sec:normalization}).}. We leverage out-of-the-box ASR solutions and experiment with Speech-to-Text services from Microsoft Azure, Google Cloud, and OpenAI Whisper. 
In general, Google and Microsoft Azure were at par in terms of supported features, such as support for streaming (real-time recognition), recognition of different dialects, spoken punctuation recognition, vocabulary customization, and price. We also considered OpenAI Whisper since it is open-source and about 70\% cheaper. We eventually chose to use the Microsoft Azure Speech-to-Text due to quality considerations (Section \ref{sec:asr_eval}).

\subsection{Normalization}
\label{sec:normalization}
The transcribed audio input may still contain noise stemming from ASR errors, speech disfluencies, uniqueness of individual elocution, ambiguous word boundaries, background noise, and lack of context, among others. 
Therefore, we introduce another stage in the pipeline to enrich the speech transcription further and get a cleaner input for the downstream comprehension model(s)\footnote{We experimented with handling these issues with the fine-tuned comprehension model (Section \ref{sec:comp-ft-details}) but found that it could not reliably fix all of the issues, further resulting in deterioration in the quality of the generated text.}. This stage comprises three sub-stages dedicated to addressing specific issues in the transcription: Speech Disfluency Filtering, Punctuation Restoration, and Grammatical Error Correction. We now describe these in more detail.

\subsubsection{Speech Disfluency Filtering}
One of the numerous issues encountered in speech-based systems pertains to the inherent fluidity of spoken language, characterized by the occurrence of errors and spontaneous self-correction. Speakers, upon recognizing their speech errors, instinctively engage in the process of rectification by means of editing, reformulating, or starting afresh. This instinctual and subconscious phenomenon is a common and integral part of spontaneous human utterance, referred to as disfluencies \cite{shriberg1994preliminaries}, and poses significant challenges to the real-world deployment of speech-based systems. 

Specifically, this part of the system focuses on detecting and removing disfluent tokens in the transcribed text and not replacing them with correct hypotheses. We formulate this as a token-level sequence tagging problem and experiment with three models. To categorize the disfluencies, we use the framework defined in \citet{shriberg1994preliminaries}, which has three categories: \textit{repetitions} (one or many words are repeated), \textit{replacements} (a disfluent word or phrase is replaced with a fluent one), and \textit{restarts} (initial utterance is completely abandoned and restarted).

Following are the details of the Disfluency Filtering models:
\begin{enumerate}
    \item \textbf{Baseline}: An off-the-shelf model for joint disfluency detection and constituency parsing \cite{jamshid-lou-johnson-2020-improving}.
    \item \textbf{\textsc{Disf-SB-QA}}: RoBERTa \cite{liu2019roberta} model, fine-tuned on two publicly available datasets: The Switchboard Corpus \cite{225858} and Disfl-QA \cite{gupta-etal-2021-disfl}.
    \item \textbf{\textsc{Disf-SB-QA-LD}}: \textsc{Disf-SB-QA} model further fine-tuned on an augmented dataset of artificial disfluencies and task-specific data using LARD \cite{passali-etal-2022-lard}.
\end{enumerate}

\subsubsection{Punctuation Restoration}
Once the disfluencies are removed, the input is still a stream of text without any punctuation or sentence segmentation. Therefore, the next step in the system is to restore punctuation (including capitalization). 
We experiment with three models that are trained to perform multi-class token classification. Specifically, there are five categories describing the respective token-level edit actions they apply:
\begin{itemize}
\item \texttt{COMMA}: Append one of [, ; : -]
\vspace{-0.1cm}
\item \texttt{PERIOD}: Append .
\vspace{-0.1cm}
\item \indent\texttt{QUESTIONMARK}: Append one of [? !]
\vspace{-0.1cm}
\item \indent\texttt{CAPITALIZATION}: Capitalize the word
\vspace{-0.1cm}
\item \indent\texttt{NONE}: No change
\end{itemize}

Following are the details of the Punctuation restoration models:
\begin{enumerate}
\item \textbf{rpunct}\footnotemark\footnotetext{\url{https://github.com/Felflare/rpunct}} is an open-source Python package for punctuation restoration, which uses a \textsc{BERT}-base model trained on Yelp reviews dataset\footnotemark\footnotetext{\url{https://www.yelp.com/dataset/challenge}}. We use this as our baseline.
\item \textbf{\textsc{Punct-Comp}}: Fine-tuned DistilBERT \cite{sanh2019distilbert} model on the same dataset as \textsc{Comp-FT} (described in Section \ref{sec:comp-ft-details}).
\item \textbf{\textsc{Punct-Comp-GEC}}: Retrained version of \textsc{Punct-Comp} model after applying grammatical corrections (Section \ref{sec:gec} ) to the dataset. 
\end{enumerate}


\subsubsection{Grammatical Error Correction (GEC)}
\label{sec:gec}
We use the GECToR system \cite{omelianchuk-etal-2020-gector} for grammatical error correction. Similar to our models for Disfluency filtering and Punctuation restoration, it is a sequence tagging model using a Transformer-based encoder. 

\subsection{Comprehension}


\begin{table*}[]
    \centering
    \small
    \begin{tabular}{p{0.45\linewidth}|c|c|c|c} 
     \toprule
     \textbf{Example} & \textbf{Input Type} & \textbf{Content Type} & \textbf{Intent Type} & \textbf{Model} \\
     \midrule
     Hey John, are you coming to the meeting later today? & Dictation & Closed-ended & Messaging & \textsc{Comp-FT}\\     
     \midrule
     Email Sam, we met with Joe today, meeting went well, follow-up with him next week. & Dictation & Open-ended & Email & \textsc{Comp-FT}\\
     \midrule     
     After more than 50 years The Eagles are heading on the road for what they say will be their "final" tour. On Thursday the legendary band announced “The Long Goodbye” tour that is set to kick off September 7 in New York. & Dictation & Closed-ended & Notes & \textsc{Comp-FT} \\
     \midrule
     Send an email to Joe. Let him know that fundraiser is a go, and it will be happening next Wednesday at 8:00. PM. & Instruction & Closed-ended & Email & \textsc{Comp-FT} \\
     \midrule
     Pick up groceries at 5 pm tomorrow. & Instruction & Closed-ended & Notes & \textsc{Comp-FT} \\
     \midrule
     Write a thoughtful birthday wish for Jim. He is one of my oldest friends. He is turning 31. Make the message witty.  & Instruction & Open-ended & Messaging & \textsc{Comp-LLM}\\
     \midrule
     Write a blog post on AI from the perspective of a 30-year-old adult. & Instruction & Open-ended & Notes & \textsc{Comp-LLM}\\
     \bottomrule
    \end{tabular}
    \caption{Different types of inputs (i.e. normalization outputs) handled by our system, along with their characteristics.}
    \label{tab:dataset}
\end{table*}

The output from the normalization step is then fed into the comprehension stage, which transforms the normalized input into a well-structured and coherent output, handling a wide variety of inputs. 
Table \ref{tab:dataset} shows the different types of inputs that the comprehension stage can handle. For example, the input can be an instruction or a dictation; an email, an instant message, or a note; open-ended or closed-ended\footnote{Closed-ended use cases are inputs which provide most of the necessary details whereas open-ended inputs require the model to fill in some details.}. Moreover, the spoken text can often be incomplete and noisy. Thus, the comprehension model enhances the quality of such text while minimizing meaning change and hallucination.

We experiment with two approaches for the comprehension stage. The first is fine-tuning a lightweight pre-trained model (called \textsc{Comp-FT}), and the second is using a pre-trained LLM out-of-the-box (called \textsc{Comp-LLM}). 


\subsubsection{\textsc{Comp-FT}}
\label{sec:comp-ft-details}
We use Pegasus \cite{zhang2020pegasus} (770M parameters), a transformer-based encoder-decoder. We limit ourselves to a small model since larger models have a higher latency, and we find that a model of this size can handle a significant portion of inputs. Since smaller models do not work well on open-ended generation, we limit it to closed-ended inputs. Model training details are present in Appendix \ref{sec:comp_ft_model}.

To fine-tune \textsc{Comp-FT}, we create a dataset containing 28k/1k/1k input-output pairs for training/validation/test sets, respectively.
First, we ask human annotators to create 10k instruction-output pairs covering the various instruction-based use cases described earlier.
Then, we create dictation-based data by removing the formatting and paraphrasing\footnote{We use fine-tuned Pegasus on Parabank \cite{hu-etal-2019-large} as our model.} the outputs from this dataset, and use the resulting text as inputs instead.

Finally, we augment the dataset by applying 25 different augmentations to deal with the issues that were either not handled or were introduced by the earlier stages of the pipeline. We build upon NL-Augmenter \cite{dhole2021nl}, an open-source library that contains 117 transformations and 23 filters for a variety of natural language tasks. A selection of the augmentations can be found in Appendix \ref{sec:augmentations}.

\subsubsection{\textsc{Comp-LLM}}

We use the \textit{gpt-3.5-turbo} model from the Azure OpenAI platform. Since this model is a chat-based model, the main challenge is to find the right prompt for all our use cases. 
Further, the text generated by it is prone to verbosity and often contains hallucinations leading to meaning change. The benefit, however, is its ability to handle open-ended inputs such as \textit{"Write a list of items to bring camping"}. Finally, it has higher latency and is more expensive to deploy. 

\subsubsection{Hybrid Approach}

Since both \textsc{Comp-FT} and \textsc{Comp-LLM} are effective at different use cases, we combine both models into a hybrid approach. Outputs requiring more open-ended generation and having low scope for sensitivity issues are passed to \textsc{Comp-LLM}, whereas shorter inputs and those which require more factual consistency are processed by \textsc{Comp-FT}. The last column in Table \ref{tab:dataset} shows which model processes the different inputs. We train a binary classifier, a fine-tuned DistilBERT \cite{sanh2019distilbert} model, to decide whether the system should use \textsc{Comp-FT} or \textsc{Comp-LLM}. This model was trained using a manually created dataset containing 1000 examples. The classifier is applied to the output text of the normalization stage. 

\section{Evaluation}


\subsection{ASR}
\label{sec:asr_eval}
In order to evaluate the quality of the various ASR systems, we collected a dataset of 1000 voice inputs by releasing the system to a small set of internal users, who were asked to use the system for their composition needs. 
Expert annotators then transcribed these voice recordings, and the ASR systems were evaluated using the standard ASR metrics of Word Error Rate (WER) and Word Recognition Rate (WRR). Table \ref{tab:asr} shows the performance comparison of these different ASR systems on this set. We found that Microsoft Azure Speech-to-Text achieved the best performance, which determined our choice of ASR system for \speakerly.

\begin{table}[]
    \centering
    \small
    \begin{tabular}{l|c|c} 
     \toprule
     \textbf{System} & \textbf{WER} (\%)$\downarrow$ & \textbf{WRR} (\%)$\uparrow$ \\ 
     \midrule
     Microsoft Azure\footnotemark[1] & \textbf{3.37} & \textbf{97.13} \\ 
     Google Speech-to-Text\footnotemark[2] & 4.55 & 96.79 \\ 
     OpenAI Whisper\footnotemark[3] & 4.43 & 96.83 \\ 
     \bottomrule
    \end{tabular}
    \caption{Performance comparison of different ASR solutions. WER indicates Word Error Rate, and WRR indicates Word Recognition Rate.}
    \label{tab:asr}
\end{table}

\footnotetext[1]{\url{https://azure.microsoft.com/en-us/products/cognitive-services/speech-to-text}}
\footnotetext[2]{\url{https://cloud.google.com/speech-to-text}}
\footnotetext[3]{\url{https://openai.com/research/whisper}}

\subsection{Speech Disfluency Filtering}
Since the Disfluency Filtering models are sequence tagging models, we use Precision/Recall/F1 as the evaluation metrics on two evaluation datasets. First is the CCPE-M dataset \cite{radlinski-etal-2019-coached}, a corpus consisting of dialogues between two paid crowd-workers using a Wizard-of-Oz based, Coached Conversational Preference Elicitation (CCPE) methodology. We also collect and annotate (via crowdsourcing) an internal dataset sourced from the transcripts of company-wide, internal Zoom meetings, which were then annotated for the disfluency filtering task by expert annotators. Table \ref{tab:disfluency} summarizes the results of the three models on the two evaluation sets. We observed that \textsc{Disf-SB-QA-LD} was the best-performing model, owing largely to the task-specific data augmentation.

\begin{table}[]
    \centering
    \small
    \begin{tabular}{l|c|c} 
     \toprule
     \textbf{System} & \textbf{CCPE-M} & \textbf{Meetings} \\ 
     \midrule
     Baseline & 59.2 / \textbf{75.3} / 66.3 & 76.5 / 51.2 / 61.3 \\
     Disf-SB-QA & \textbf{83.5} / 55.0 / 66.3 &  87.4 / 82.2 / 84.7  \\
     Disf-SB-QA-LD & 78.7 / 68.4 / \textbf{73.2} & \textbf{97.3} / \textbf{89.5} / \textbf{93.2} \\
\bottomrule
    \end{tabular}
    \caption{Performance comparison of different Disfluency Filtering models (Precision / Recall / F1).}
    \vspace{-0.4cm}
    \label{tab:disfluency}
\end{table}

\subsection{Punctuation Restoration}
Since the Punctuation Restoration models are also sequence tagging models, we evaluate them using Precision/Recall/F1 metrics on the same test set as the COMP-FT model (Section \ref{sec:comp-ft-details}). Table \ref{tab:punctuation} details the results of the three models on the test set for all the punctuation label groups. We also report metrics for sentence boundary detection, which is a combination of the \texttt{PERIOD} and \texttt{QUESTIONMARK} labels. We observe that \textsc{Punct-Comp-GEC} was the best-performing model in most categories.

\begin{table*}[]
    \centering
    \small
    \begin{tabular}{l|c|c|c|c} 
     \toprule
     \textbf{Model} & \textbf{Sentence} & \textbf{Comma} & \textbf{Period} & \textbf{Question} \\
     \midrule
     rpunct & 91.0 / 88.3 / 89.6 & 72.5 / 42.8 / 53.8 & 83.0 / 91.4 / 87.0 & 78.4 / 81.0 / 79.7 \\
     Punct-Comp & \textbf{94.3} / \textbf{92.7} / \textbf{93.5} & 80.2 / 78.5 / 79.3 & 90.6 / 87.6 / 89.1 &  \textbf{89.0} / 83.5 / \textbf{86.2} \\
     Punct-Comp-GEC & \textbf{94.3} / 92.1 / 93.2	& \textbf{80.7} / \textbf{93.9} / \textbf{86.8} & \textbf{93.9} / \textbf{92.5} / \textbf{93.2} & 78.6 / \textbf{92.6} / 85.0 \\
     \bottomrule
    \end{tabular}
    \caption{Performance comparison of different Punctuation Restoration models (Precision / Recall / F1).}
    \label{tab:punctuation}
\end{table*}

\subsection{Comprehension}
\label{sec:comprehension_automatic}

\begin{table*}[]
    \centering
    \small
    \begin{tabular}{l|c|c|c|c} 
     \toprule
     \textbf{System} & \textbf{Fluency} (\%)$\uparrow$ & \textbf{Coherence} (\%)$\uparrow$ & \textbf{Naturalness} (\%)$\uparrow$ & \textbf{Coverage} (\%)$\uparrow$ \\ 
     \midrule
     \textsc{Comp-FT} & 66.49 (0.61/83.47) & \textbf{88.47} (0.31/83.35) & \textbf{76.93} (0.43/78.80) & \textbf{83.31} (0.51/85.32)  \\ 
     \textsc{Comp-LLM} & \textbf{68.93} (0.57/82.40) & 86.02 (0.35/82.98) & 75.98 (0.45/78.82) & 68.25 (0.55/80.56)  \\ 
     \bottomrule
    \end{tabular}
    \caption{Human Evaluation of different comprehension models on Fluency, Coherence, Naturalness, and Coverage. The number in brackets shows Cohen's $\kappa$ and inter-annotator agreement scores, respectively.}
    \label{tab:comprehension_human}
\end{table*}

For \textsc{Comp-FT}, we evaluate various models between 240M and 1.3B parameters on our test set (Section \ref{sec:comp-ft-details}) using BLEU \cite{papineni-etal-2002-bleu}, ROUGE \cite{lin-2004-rouge}, METEOR \cite{banerjee-lavie-2005-meteor} and BLEURT \cite{sellam-etal-2020-bleurt} as evaluation metrics. The Pegasus \cite{zhang2020pegasus} model outperforms the other models (Table \ref{tab:comprehension_auto}). However, we find that automated metrics were neither suitable nor reliable for evaluation, as they largely focus on n-gram-based overlap with references. 
Thus, we use human evaluations to measure the quality of our comprehension models. 

We conduct extensive human annotation studies to gather insight into the quality of the output generated by the comprehension models. First, we compare \textsc{Comp-FT} and \textsc{Comp-LLM} on various closed-ended composition scenarios using 1200 examples. We restrict this dataset to closed-ended use cases since \textsc{Comp-FT} does not work well for open-ended use cases.  
For each example, we ask seven annotators to provide a binary judgment on fluency, coherence, naturalness, and coverage (descriptions to annotators provided in Appendix \ref{sec:comp_human_eval_guidelines}) and decide the final judgment by majority voting. 
We also measure inter-annotator agreement using a simple percent agreement, as well as Cohen's $\kappa$ \cite{mchugh2012interrater}.

Table \ref{tab:comprehension_human} shows the human evaluation results for the two models on the four metrics and the corresponding inter-annotator agreement scores. 
We find that outputs generated by \textsc{Comp-LLM} are more fluent than those from \textsc{Comp-FT}. This result is expected since LLMs are known to generate highly fluent text. 
Further, outputs generated by \textsc{Comp-FT} are marginally better than those from \textsc{Comp-LLM} on coherence and naturalness. 
Finally, we find that outputs generated by \textsc{Comp-LLM} have much more meaning-change than those from \textsc{Comp-FT}, highlighting a known problem of hallucination in LLMs. 
Overall, we find that in closed-ended inputs, the text generated by \textsc{Comp-FT} is overall of higher quality than that generated by \textsc{Comp-LLM}.

Table \ref{tab:comprehension_human} also shows that Cohen's $\kappa$ scores were higher for both models on Fluency and Coverage, indicating that annotators were more aligned on these criteria than they were on Coherence and Naturalness. 
This confirms our understanding that grammar and the presence or absence of information are more objective, whereas Coherence and Naturalness are more subjective and may vary based on context (for example, a short message may be unnatural but perfectly acceptable as a quick reply). Even though these categories had lower $\kappa$ scores, they are still in a range that is considered fair agreement. 

\subsubsection{Sensitivity Evaluations}
Current text generation systems have been shown to contain bias and behave differently to sensitive text \cite{bender2021dangers, welbl-etal-2021-challenges-detoxifying, hovy2021five}. Therefore, we conduct an iterative sensitivity review of our end-to-end pipeline. We prepare a dataset of 800 sensitive examples to test the generation quality on offensive and non-inclusive language, bias, meaning change, and sensitive domains (such as medical advice and self-harm).
We reviewed the generated outputs for the sensitive inputs and after manual reviewing, made the following changes to mitigate the identified risks:

\begin{enumerate}
    \item Apply dictionary-based filtering for offensive words and a sensitivity classifier\footnote{DistilBERT \cite{sanh2019distilbert} trained on DeTexD dataset \cite{chernodub-etal-2023-detexd}} after both the normalization and comprehension stages.
    \item Retrain \textsc{Comp-FT} on an improved dataset containing examples to handle sensitive text better, improved co-reference resolution, and diversity-based augmentations. For \textsc{Comp-LLM}, we evaluate prompts on their ability to handle sensitive text.
    \item Adjust the classifier of the hybrid model to send more sensitive data to \textsc{Comp-FT} instead of \textsc{Comp-LLM}.
\end{enumerate}

Overall, we find that \textsc{Comp-FT} is much better at handling sensitive text compared to \textsc{Comp-LLM}.

\subsection{Inference}

We deploy our service on Amazon ECS using the \texttt{g5.2xlarge} instances. To increase the throughput while reducing overall latency, we enable our service to scale horizontally as well as run multiple inference workers per instance.
We conduct load testing to evaluate the infrastructure costs required for deploying the system. We find that we can successfully serve a constant traffic of 1 request per second using the \textsc{Comp-FT} model on a single \texttt{g5.2xlarge} instance while maintaining a p90 latency of 3 seconds. To achieve the same latency and throughput requirements for \textsc{Comp-LLM}, we need to scale the number of instances to 30. With a hybrid system that routes the request to either \textsc{Comp-FT} or \textsc{Comp-LLM}, we can reduce the number of instances to 10.

\section{Conclusion}

In this paper, we presented \speakerly, a real-time voice-based writing assistant for text composition. It provides a low barrier to entry into the writing process, where a user can interact naturally, either using dictation, instructions, or unstructured thoughts. 
In turn, it generates a high-quality first draft with low latency, thus, providing them with a simple and efficient way to articulate their thoughts into ready-to-send emails, messages, or notes. We present comprehensive technical details of the different stages of the pipeline and experiments which guided our decisions while deploying the system to our users.

\section*{Limitations}
While we design \speakerly to handle the various challenges that can occur in real-world spoken input, there are instances where the system can generate output that does not reflect what the user wanted to say or generate sensitive text. In such cases, the user can either ask the system to regenerate the output, speak again, or manually edit the generated output. Since manually editing the system can be tedious, we plan to integrate a text editing step in the pipeline. Furthermore, our system currently cannot generate very long outputs (greater than 512 tokens). Currently, for most open-ended inputs, we rely on an external LLM, which can be costly and have high latency. Moving forward, we intend to look at other smaller models that can generate high-quality outputs for such texts. Lastly, since we use external ASR systems, which can be limited in their ability to deal with different accents, our system's ability can be limited by it (even though we do have augmentations to mimic such inputs). Finally, we only tested this system for English.

\section*{Ethics Statement}
During the data annotation and model evaluation processes, all human evaluators' identities were anonymized to respect their privacy rights. All human evaluators received a fair wage higher than the minimum wage based on the number of data points they evaluated.

Although we implement ways to mitigate risks associated with sensitive texts, there can still be instances where the model can generate some sensitive output or cause meaning change and hallucination, especially for open-ended inputs. We do give users options to give feedback and report such issues, which we plan to keep improving the system on (using signals such as social factors, for instance, \cite{kulkarni2023writing}). 

\section*{Acknowledgements}
We thank the anonymous reviewers for their detailed and helpful reviews. We also appreciate the outstanding contributions of our colleagues in the Grammarly Intelligence Org who 
helped us at various stages in the project.
Finally, we would like to thank Gaurav Sahu, Vivek Kulkarni, and Max Gubin, who provided helpful feedback on the manuscript.

\bibliography{anthology,custom}

\begin{thebibliography}{55}
\expandafter\ifx\csname natexlab\endcsname\relax\def\natexlab#1{#1}\fi

\bibitem[{Banerjee and Lavie(2005)}]{banerjee-lavie-2005-meteor}
Satanjeev Banerjee and Alon Lavie. 2005.
\newblock \href {https://aclanthology.org/W05-0909} {{METEOR}: An automatic
  metric for {MT} evaluation with improved correlation with human judgments}.
\newblock In \emph{Proceedings of the {ACL} Workshop on Intrinsic and Extrinsic
  Evaluation Measures for Machine Translation and/or Summarization}, pages
  65--72, Ann Arbor, Michigan. Association for Computational Linguistics.

\bibitem[{Bender et~al.(2021)Bender, Gebru, McMillan-Major, and
  Shmitchell}]{bender2021dangers}
Emily~M Bender, Timnit Gebru, Angelina McMillan-Major, and Shmargaret
  Shmitchell. 2021.
\newblock On the dangers of stochastic parrots: Can language models be too big?
\newblock In \emph{Proceedings of the 2021 ACM conference on fairness,
  accountability, and transparency}, pages 610--623.

\bibitem[{Black et~al.(2021)Black, Gao, Wang, Leahy, and
  Biderman}]{black2021gpt}
Sid Black, Leo Gao, Phil Wang, Connor Leahy, and Stella Biderman. 2021.
\newblock \href {https://doi.org/10.5281/zenodo.5297715} {{GPT-Neo: Large Scale
  Autoregressive Language Modeling with Mesh-Tensorflow}}.

\bibitem[{Chakrabarty et~al.(2022)Chakrabarty, Padmakumar, and
  He}]{chakrabarty-etal-2022-help}
Tuhin Chakrabarty, Vishakh Padmakumar, and He~He. 2022.
\newblock \href {https://doi.org/10.18653/v1/2022.emnlp-main.460} {Help me
  write a poem: Instruction tuning as a vehicle for collaborative poetry
  writing}.
\newblock In \emph{Proceedings of the 2022 Conference on Empirical Methods in
  Natural Language Processing}, pages 6848--6863, Abu Dhabi, United Arab
  Emirates. Association for Computational Linguistics.

\bibitem[{Clark et~al.(2018)Clark, Ross, Tan, Ji, and
  Smith}]{clark2018creative}
Elizabeth Clark, Anne~Spencer Ross, Chenhao Tan, Yangfeng Ji, and Noah~A Smith.
  2018.
\newblock Creative writing with a machine in the loop: Case studies on slogans
  and stories.
\newblock In \emph{23rd International Conference on Intelligent User
  Interfaces}, pages 329--340.

\bibitem[{Coenen et~al.(2021)Coenen, Davis, Ippolito, Reif, and
  Yuan}]{coenen2021wordcraft}
Andy Coenen, Luke Davis, Daphne Ippolito, Emily Reif, and Ann Yuan. 2021.
\newblock Wordcraft: a human-ai collaborative editor for story writing.
\newblock \emph{arXiv preprint arXiv:2107.07430}.

\bibitem[{Cohen and Oviatt(1995)}]{doi:10.1073/pnas.92.22.9921}
P~R Cohen and S~L Oviatt. 1995.
\newblock \href {https://doi.org/10.1073/pnas.92.22.9921} {The role of voice
  input for human-machine communication.}
\newblock \emph{Proceedings of the National Academy of Sciences},
  92(22):9921--9927.

\bibitem[{Dhole et~al.(2021)Dhole, Gangal, Gehrmann, Gupta, Li, Mahamood,
  Mahendiran, Mille, Shrivastava, Tan et~al.}]{dhole2021nl}
Kaustubh~D Dhole, Varun Gangal, Sebastian Gehrmann, Aadesh Gupta, Zhenhao Li,
  Saad Mahamood, Abinaya Mahendiran, Simon Mille, Ashish Shrivastava, Samson
  Tan, et~al. 2021.
\newblock Nl-augmenter: A framework for task-sensitive natural language
  augmentation.
\newblock \emph{arXiv preprint arXiv:2112.02721}.

\bibitem[{Du et~al.(2022)Du, Kim, Raheja, Kumar, and Kang}]{du-etal-2022-read}
Wanyu Du, Zae~Myung Kim, Vipul Raheja, Dhruv Kumar, and Dongyeop Kang. 2022.
\newblock \href {https://doi.org/10.18653/v1/2022.in2writing-1.14} {Read,
  revise, repeat: A system demonstration for human-in-the-loop iterative text
  revision}.
\newblock In \emph{Proceedings of the First Workshop on Intelligent and
  Interactive Writing Assistants (In2Writing 2022)}, pages 96--108, Dublin,
  Ireland. Association for Computational Linguistics.

\bibitem[{Fan et~al.(2021)Fan, Xu, Yu, and Shi}]{fan2021just}
Jiayue Fan, Chenning Xu, Chun Yu, and Yuanchun Shi. 2021.
\newblock Just speak it: Minimize cognitive load for eyes-free text editing
  with a smart voice assistant.
\newblock In \emph{The 34th Annual ACM Symposium on User Interface Software and
  Technology}, pages 910--921.

\bibitem[{Flower and Hayes(1981)}]{f508427a-e4c0-3d6a-8abf-03a5d21ec6c4}
Linda Flower and John~R. Hayes. 1981.
\newblock \href {http://www.jstor.org/stable/356600} {A cognitive process
  theory of writing}.
\newblock \emph{College Composition and Communication}, 32(4):365--387.

\bibitem[{Gero et~al.(2022{\natexlab{a}})Gero, Calderwood, Li, and
  Chilton}]{gero-etal-2022-design}
Katy Gero, Alex Calderwood, Charlotte Li, and Lydia Chilton.
  2022{\natexlab{a}}.
\newblock \href {https://doi.org/10.18653/v1/2022.in2writing-1.2} {A design
  space for writing support tools using a cognitive process model of writing}.
\newblock In \emph{Proceedings of the First Workshop on Intelligent and
  Interactive Writing Assistants (In2Writing 2022)}, pages 11--24, Dublin,
  Ireland. Association for Computational Linguistics.

\bibitem[{Gero et~al.(2022{\natexlab{b}})Gero, Liu, and
  Chilton}]{gero-etal-2022-sparks}
Katy Gero, Vivian Liu, and Lydia Chilton. 2022{\natexlab{b}}.
\newblock \href {https://doi.org/10.18653/v1/2022.in2writing-1.12} {Sparks:
  Inspiration for science writing using language models}.
\newblock In \emph{Proceedings of the First Workshop on Intelligent and
  Interactive Writing Assistants (In2Writing 2022)}, pages 83--84, Dublin,
  Ireland. Association for Computational Linguistics.

\bibitem[{Gero and Chilton(2019)}]{gero2019metaphoria}
Katy~Ilonka Gero and Lydia~B Chilton. 2019.
\newblock Metaphoria: An algorithmic companion for metaphor creation.
\newblock In \emph{Proceedings of the 2019 CHI conference on human factors in
  computing systems}, pages 1--12.

\bibitem[{Ghosh(2020)}]{ghosh2020voice}
Debjyoti Ghosh. 2020.
\newblock \emph{Voice-based Interactions for Editing Text On The Go}.
\newblock Ph.D. thesis, National University of Singapore (Singapore).

\bibitem[{Godfrey et~al.(1992)Godfrey, Holliman, and McDaniel}]{225858}
J.J. Godfrey, E.C. Holliman, and J.~McDaniel. 1992.
\newblock \href {https://doi.org/10.1109/ICASSP.1992.225858} {Switchboard:
  telephone speech corpus for research and development}.
\newblock In \emph{[Proceedings] ICASSP-92: 1992 IEEE International Conference
  on Acoustics, Speech, and Signal Processing}, volume~1, pages 517--520 vol.1.

\bibitem[{Gupta et~al.(2021)Gupta, Xu, Upadhyay, Yang, and
  Faruqui}]{gupta-etal-2021-disfl}
Aditya Gupta, Jiacheng Xu, Shyam Upadhyay, Diyi Yang, and Manaal Faruqui. 2021.
\newblock \href {https://doi.org/10.18653/v1/2021.findings-acl.293}
  {Disfl-{QA}: A benchmark dataset for understanding disfluencies in question
  answering}.
\newblock In \emph{Findings of the Association for Computational Linguistics:
  ACL-IJCNLP 2021}, pages 3309--3319, Online. Association for Computational
  Linguistics.

\bibitem[{Hovy and Prabhumoye(2021)}]{hovy2021five}
Dirk Hovy and Shrimai Prabhumoye. 2021.
\newblock Five sources of bias in natural language processing.
\newblock \emph{Language and Linguistics Compass}, 15(8):e12432.

\bibitem[{Hu et~al.(2019)Hu, Singh, Holzenberger, Post, and
  Van~Durme}]{hu-etal-2019-large}
J.~Edward Hu, Abhinav Singh, Nils Holzenberger, Matt Post, and Benjamin
  Van~Durme. 2019.
\newblock \href {https://doi.org/10.18653/v1/K19-1005} {Large-scale, diverse,
  paraphrastic bitexts via sampling and clustering}.
\newblock In \emph{Proceedings of the 23rd Conference on Computational Natural
  Language Learning (CoNLL)}, pages 44--54, Hong Kong, China. Association for
  Computational Linguistics.

\bibitem[{Hui et~al.(2018)Hui, Gergle, and Gerber}]{hui2018introassist}
Julie~S Hui, Darren Gergle, and Elizabeth~M Gerber. 2018.
\newblock Introassist: A tool to support writing introductory help requests.
\newblock In \emph{Proceedings of the 2018 CHI Conference on Human Factors in
  Computing Systems}, pages 1--13.

\bibitem[{Jamshid~Lou and Johnson(2020)}]{jamshid-lou-johnson-2020-improving}
Paria Jamshid~Lou and Mark Johnson. 2020.
\newblock \href {https://doi.org/10.18653/v1/2020.acl-main.346} {Improving
  disfluency detection by self-training a self-attentive model}.
\newblock In \emph{Proceedings of the 58th Annual Meeting of the Association
  for Computational Linguistics}, pages 3754--3763, Online. Association for
  Computational Linguistics.

\bibitem[{Kamm(1995)}]{doi:10.1073/pnas.92.22.10031}
C~Kamm. 1995.
\newblock \href {https://doi.org/10.1073/pnas.92.22.10031} {User interfaces for
  voice applications.}
\newblock \emph{Proceedings of the National Academy of Sciences},
  92(22):10031--10037.

\bibitem[{Kim et~al.(2022)Kim, Du, Raheja, Kumar, and
  Kang}]{kim-etal-2022-improving}
Zae~Myung Kim, Wanyu Du, Vipul Raheja, Dhruv Kumar, and Dongyeop Kang. 2022.
\newblock \href {https://doi.org/10.18653/v1/2022.emnlp-main.678} {Improving
  iterative text revision by learning where to edit from other revision tasks}.
\newblock In \emph{Proceedings of the 2022 Conference on Empirical Methods in
  Natural Language Processing}, pages 9986--9999, Abu Dhabi, United Arab
  Emirates. Association for Computational Linguistics.

\bibitem[{Kulkarni and Raheja(2023)}]{kulkarni2023writing}
Vivek Kulkarni and Vipul Raheja. 2023.
\newblock \href {http://arxiv.org/abs/2303.16275} {Writing assistants should
  model social factors of language}.

\bibitem[{Lewis et~al.(2020)Lewis, Liu, Goyal, Ghazvininejad, Mohamed, Levy,
  Stoyanov, and Zettlemoyer}]{lewis-etal-2020-bart}
Mike Lewis, Yinhan Liu, Naman Goyal, Marjan Ghazvininejad, Abdelrahman Mohamed,
  Omer Levy, Veselin Stoyanov, and Luke Zettlemoyer. 2020.
\newblock \href {https://doi.org/10.18653/v1/2020.acl-main.703} {{BART}:
  Denoising sequence-to-sequence pre-training for natural language generation,
  translation, and comprehension}.
\newblock In \emph{Proceedings of the 58th Annual Meeting of the Association
  for Computational Linguistics}, pages 7871--7880, Online. Association for
  Computational Linguistics.

\bibitem[{Lin(2004)}]{lin-2004-rouge}
Chin-Yew Lin. 2004.
\newblock \href {https://aclanthology.org/W04-1013} {{ROUGE}: A package for
  automatic evaluation of summaries}.
\newblock In \emph{Text Summarization Branches Out}, pages 74--81, Barcelona,
  Spain. Association for Computational Linguistics.

\bibitem[{Liu et~al.(2019)Liu, Ott, Goyal, Du, Joshi, Chen, Levy, Lewis,
  Zettlemoyer, and Stoyanov}]{liu2019roberta}
Yinhan Liu, Myle Ott, Naman Goyal, Jingfei Du, Mandar Joshi, Danqi Chen, Omer
  Levy, Mike Lewis, Luke Zettlemoyer, and Veselin Stoyanov. 2019.
\newblock Roberta: A robustly optimized bert pretraining approach.
\newblock \emph{arXiv preprint arXiv:1907.11692}.

\bibitem[{Luo et~al.(2021)Luo, Kim, Lee, Hassan, and Choe}]{luo2021foodscrap}
Yuhan Luo, Young-Ho Kim, Bongshin Lee, Naeemul Hassan, and Eun~Kyoung Choe.
  2021.
\newblock Foodscrap: Promoting rich data capture and reflective food journaling
  through speech input.
\newblock In \emph{Designing Interactive Systems Conference 2021}, pages
  606--618.

\bibitem[{Luo et~al.(2020)Luo, Lee, and Choe}]{luo2020tandemtrack}
Yuhan Luo, Bongshin Lee, and Eun~Kyoung Choe. 2020.
\newblock Tandemtrack: shaping consistent exercise experience by complementing
  a mobile app with a smart speaker.
\newblock In \emph{Proceedings of the 2020 CHI Conference on Human Factors in
  Computing Systems}, pages 1--13.

\bibitem[{Mallinson et~al.(2022)Mallinson, Adamek, Malmi, and
  Severyn}]{mallinson-etal-2022-edit5}
Jonathan Mallinson, Jakub Adamek, Eric Malmi, and Aliaksei Severyn. 2022.
\newblock \href {https://doi.org/10.18653/v1/2022.findings-emnlp.156}
  {{E}di{T}5: Semi-autoregressive text editing with t5 warm-start}.
\newblock In \emph{Findings of the Association for Computational Linguistics:
  EMNLP 2022}, pages 2126--2138, Abu Dhabi, United Arab Emirates. Association
  for Computational Linguistics.

\bibitem[{McHugh(2012)}]{mchugh2012interrater}
Mary~L McHugh. 2012.
\newblock Interrater reliability: the kappa statistic.
\newblock \emph{Biochemia medica}, 22(3):276--282.

\bibitem[{Omelianchuk et~al.(2020)Omelianchuk, Atrasevych, Chernodub, and
  Skurzhanskyi}]{omelianchuk-etal-2020-gector}
Kostiantyn Omelianchuk, Vitaliy Atrasevych, Artem Chernodub, and Oleksandr
  Skurzhanskyi. 2020.
\newblock \href {https://doi.org/10.18653/v1/2020.bea-1.16} {{GECT}o{R} {--}
  grammatical error correction: Tag, not rewrite}.
\newblock In \emph{Proceedings of the Fifteenth Workshop on Innovative Use of
  NLP for Building Educational Applications}, pages 163--170, Seattle, WA, USA
  → Online. Association for Computational Linguistics.

\bibitem[{Papineni et~al.(2002)Papineni, Roukos, Ward, and
  Zhu}]{papineni-etal-2002-bleu}
Kishore Papineni, Salim Roukos, Todd Ward, and Wei-Jing Zhu. 2002.
\newblock \href {https://doi.org/10.3115/1073083.1073135} {{B}leu: a method for
  automatic evaluation of machine translation}.
\newblock In \emph{Proceedings of the 40th Annual Meeting of the Association
  for Computational Linguistics}, pages 311--318, Philadelphia, Pennsylvania,
  USA. Association for Computational Linguistics.

\bibitem[{Passali et~al.(2022)Passali, Mavropoulos, Tsoumakas, Meditskos, and
  Vrochidis}]{passali-etal-2022-lard}
Tatiana Passali, Thanassis Mavropoulos, Grigorios Tsoumakas, Georgios
  Meditskos, and Stefanos Vrochidis. 2022.
\newblock \href {https://aclanthology.org/2022.lrec-1.249} {{LARD}: Large-scale
  artificial disfluency generation}.
\newblock In \emph{Proceedings of the Thirteenth Language Resources and
  Evaluation Conference}, pages 2327--2336, Marseille, France. European
  Language Resources Association.

\bibitem[{Peng et~al.(2020)Peng, Guo, Tsang, and Ma}]{peng2020exploring}
Zhenhui Peng, Qingyu Guo, Ka~Wing Tsang, and Xiaojuan Ma. 2020.
\newblock Exploring the effects of technological writing assistance for support
  providers in online mental health community.
\newblock In \emph{Proceedings of the 2020 CHI Conference on Human Factors in
  Computing Systems}, pages 1--15.

\bibitem[{Porcheron et~al.(2018)Porcheron, Fischer, Reeves, and
  Sharples}]{10.1145/3173574.3174214}
Martin Porcheron, Joel~E. Fischer, Stuart Reeves, and Sarah Sharples. 2018.
\newblock \href {https://doi.org/10.1145/3173574.3174214} {Voice interfaces in
  everyday life}.
\newblock In \emph{Proceedings of the 2018 CHI Conference on Human Factors in
  Computing Systems}, CHI '18, page 1–12, New York, NY, USA. Association for
  Computing Machinery.

\bibitem[{Qi et~al.(2020)Qi, Yan, Gong, Liu, Duan, Chen, Zhang, and
  Zhou}]{qi-etal-2020-prophetnet}
Weizhen Qi, Yu~Yan, Yeyun Gong, Dayiheng Liu, Nan Duan, Jiusheng Chen, Ruofei
  Zhang, and Ming Zhou. 2020.
\newblock \href {https://doi.org/10.18653/v1/2020.findings-emnlp.217}
  {{P}rophet{N}et: Predicting future n-gram for
  sequence-to-{S}equence{P}re-training}.
\newblock In \emph{Findings of the Association for Computational Linguistics:
  EMNLP 2020}, pages 2401--2410, Online. Association for Computational
  Linguistics.

\bibitem[{Radford et~al.(2019)Radford, Wu, Child, Luan, Amodei, Sutskever
  et~al.}]{radford2019language}
Alec Radford, Jeffrey Wu, Rewon Child, David Luan, Dario Amodei, Ilya
  Sutskever, et~al. 2019.
\newblock Language models are unsupervised multitask learners.
\newblock \emph{OpenAI blog}, 1(8):9.

\bibitem[{Radlinski et~al.(2019)Radlinski, Balog, Byrne, and
  Krishnamoorthi}]{radlinski-etal-2019-coached}
Filip Radlinski, Krisztian Balog, Bill Byrne, and Karthik Krishnamoorthi. 2019.
\newblock \href {https://doi.org/10.18653/v1/W19-5941} {Coached conversational
  preference elicitation: A case study in understanding movie preferences}.
\newblock In \emph{Proceedings of the 20th Annual SIGdial Meeting on Discourse
  and Dialogue}, pages 353--360, Stockholm, Sweden. Association for
  Computational Linguistics.

\bibitem[{Raffel et~al.(2020)Raffel, Shazeer, Roberts, Lee, Narang, Matena,
  Zhou, Li, and Liu}]{2020t5}
Colin Raffel, Noam Shazeer, Adam Roberts, Katherine Lee, Sharan Narang, Michael
  Matena, Yanqi Zhou, Wei Li, and Peter~J. Liu. 2020.
\newblock \href {http://jmlr.org/papers/v21/20-074.html} {Exploring the limits
  of transfer learning with a unified text-to-text transformer}.
\newblock \emph{Journal of Machine Learning Research}, 21(140):1--67.

\bibitem[{Raheja et~al.(2023)Raheja, Kumar, Koo, and Kang}]{raheja2023coedit}
Vipul Raheja, Dhruv Kumar, Ryan Koo, and Dongyeop Kang. 2023.
\newblock Coedit: Text editing by task-specific instruction tuning.
\newblock \emph{arXiv preprint arXiv:2305.09857}.

\bibitem[{Ruan et~al.(2018)Ruan, Wobbrock, Liou, Ng, and
  Landay}]{10.1145/3161187}
Sherry Ruan, Jacob~O. Wobbrock, Kenny Liou, Andrew Ng, and James~A. Landay.
  2018.
\newblock \href {https://doi.org/10.1145/3161187} {Comparing speech and
  keyboard text entry for short messages in two languages on touchscreen
  phones}.
\newblock \emph{Proc. ACM Interact. Mob. Wearable Ubiquitous Technol.}, 1(4).

\bibitem[{Sanh et~al.(2019)Sanh, Debut, Chaumond, and
  Wolf}]{sanh2019distilbert}
Victor Sanh, Lysandre Debut, Julien Chaumond, and Thomas Wolf. 2019.
\newblock Distilbert, a distilled version of bert: smaller, faster, cheaper and
  lighter.
\newblock \emph{arXiv preprint arXiv:1910.01108}.

\bibitem[{Schalkwyk et~al.(2010)Schalkwyk, Beeferman, Beaufays, Byrne, Chelba,
  Cohen, Kamvar, and Strope}]{schalkwyk2010your}
Johan Schalkwyk, Doug Beeferman, Fran{\c{c}}oise Beaufays, Bill Byrne, Ciprian
  Chelba, Mike Cohen, Maryam Kamvar, and Brian Strope. 2010.
\newblock “your word is my command”: Google search by voice: A case study.
\newblock \emph{Advances in Speech Recognition: Mobile Environments, Call
  Centers and Clinics}, pages 61--90.

\bibitem[{Schick et~al.(2023)Schick, Yu, Jiang, Petroni, Lewis, Izacard, You,
  Nalmpantis, Grave, and Riedel}]{schick2023peer}
Timo Schick, Jane~A. Yu, Zhengbao Jiang, Fabio Petroni, Patrick Lewis, Gautier
  Izacard, Qingfei You, Christoforos Nalmpantis, Edouard Grave, and Sebastian
  Riedel. 2023.
\newblock \href {https://openreview.net/forum?id=KbYevcLjnc} {{PEER}: A
  collaborative language model}.
\newblock In \emph{The Eleventh International Conference on Learning
  Representations}.

\bibitem[{Sellam et~al.(2020)Sellam, Das, and Parikh}]{sellam-etal-2020-bleurt}
Thibault Sellam, Dipanjan Das, and Ankur Parikh. 2020.
\newblock \href {https://doi.org/10.18653/v1/2020.acl-main.704} {{BLEURT}:
  Learning robust metrics for text generation}.
\newblock In \emph{Proceedings of the 58th Annual Meeting of the Association
  for Computational Linguistics}, pages 7881--7892, Online. Association for
  Computational Linguistics.

\bibitem[{Shoeybi et~al.(2019)Shoeybi, Patwary, Puri, LeGresley, Casper, and
  Catanzaro}]{shoeybi2019megatron}
Mohammad Shoeybi, Mostofa Patwary, Raul Puri, Patrick LeGresley, Jared Casper,
  and Bryan Catanzaro. 2019.
\newblock Megatron-lm: Training multi-billion parameter language models using
  model parallelism.
\newblock \emph{arXiv preprint arXiv:1909.08053}.

\bibitem[{Shriberg(1994)}]{shriberg1994preliminaries}
E.E. Shriberg. 1994.
\newblock \href {https://books.google.ca/books?id=J2dMAQAAMAAJ}
  {\emph{Preliminaries to a Theory of Speech Disfluencies}}.
\newblock University of California, Berkeley.

\bibitem[{Srinivasan et~al.(2020)Srinivasan, Lee, Henry~Riche, Drucker, and
  Hinckley}]{srinivasan2020inchorus}
Arjun Srinivasan, Bongshin Lee, Nathalie Henry~Riche, Steven~M Drucker, and Ken
  Hinckley. 2020.
\newblock Inchorus: Designing consistent multimodal interactions for data
  visualization on tablet devices.
\newblock In \emph{Proceedings of the 2020 CHI conference on human factors in
  computing systems}, pages 1--13.

\bibitem[{Stifelman et~al.(1993)Stifelman, Arons, Schmandt, and
  Hulteen}]{stifelman1993voicenotes}
Lisa~J Stifelman, Barry Arons, Chris Schmandt, and Eric~A Hulteen. 1993.
\newblock Voicenotes: A speech interface for a hand-held voice notetaker.
\newblock In \emph{Proceedings of the INTERACT'93 and CHI'93 conference on
  Human factors in computing systems}, pages 179--186.

\bibitem[{Welbl et~al.(2021)Welbl, Glaese, Uesato, Dathathri, Mellor,
  Hendricks, Anderson, Kohli, Coppin, and
  Huang}]{welbl-etal-2021-challenges-detoxifying}
Johannes Welbl, Amelia Glaese, Jonathan Uesato, Sumanth Dathathri, John Mellor,
  Lisa~Anne Hendricks, Kirsty Anderson, Pushmeet Kohli, Ben Coppin, and Po-Sen
  Huang. 2021.
\newblock \href {https://doi.org/10.18653/v1/2021.findings-emnlp.210}
  {Challenges in detoxifying language models}.
\newblock In \emph{Findings of the Association for Computational Linguistics:
  EMNLP 2021}, pages 2447--2469, Punta Cana, Dominican Republic. Association
  for Computational Linguistics.

\bibitem[{Williams(1998)}]{williams1998guidelines}
James~R Williams. 1998.
\newblock Guidelines for the use of multimedia in instruction.
\newblock In \emph{Proceedings of the Human Factors and Ergonomics Society
  Annual Meeting}, volume~42, pages 1447--1451. SAGE Publications Sage CA: Los
  Angeles, CA.

\bibitem[{Yavnyi et~al.(2023)Yavnyi, Sliusarenko, Razzaghi, Nahorna, Mo,
  Hovakimyan, and Chernodub}]{chernodub-etal-2023-detexd}
Serhii Yavnyi, Oleksii Sliusarenko, Jade Razzaghi, Olena Nahorna, Yichen Mo,
  Knar Hovakimyan, and Artem Chernodub. 2023.
\newblock \href {https://doi.org/10.18653/v1/2023.woah-1.2} {{D}e{T}ex{D}: A
  benchmark dataset for delicate text detection}.
\newblock In \emph{The 7th Workshop on Online Abuse and Harms (WOAH)}, pages
  14--28, Toronto, Canada. Association for Computational Linguistics.

\bibitem[{Zaheer et~al.(2020)Zaheer, Guruganesh, Dubey, Ainslie, Alberti,
  Ontanon, Pham, Ravula, Wang, Yang et~al.}]{zaheer2020big}
Manzil Zaheer, Guru Guruganesh, Kumar~Avinava Dubey, Joshua Ainslie, Chris
  Alberti, Santiago Ontanon, Philip Pham, Anirudh Ravula, Qifan Wang, Li~Yang,
  et~al. 2020.
\newblock Big bird: Transformers for longer sequences.
\newblock \emph{Advances in neural information processing systems},
  33:17283--17297.

\bibitem[{Zhang et~al.(2020)Zhang, Zhao, Saleh, and Liu}]{zhang2020pegasus}
Jingqing Zhang, Yao Zhao, Mohammad Saleh, and Peter Liu. 2020.
\newblock Pegasus: Pre-training with extracted gap-sentences for abstractive
  summarization.
\newblock In \emph{International Conference on Machine Learning}, pages
  11328--11339. PMLR.

\end{thebibliography}
\bibliographystyle{acl_natbib}

\appendix


\section{\textsc{Comp-FT} Training Details}
\label{sec:comp_ft_model}
For \textsc{Comp-FT}, we evaluate various models between 240M and 1.3B parameters on our test set containing 1k examples. Table \ref{tab:comprehension_auto} shows the models, their respective parameters and their performance using BLEU, ROUGE-1, ROUGE-2, ROUGE-L, METEOR, and BLEURT metrics. We find that overall the Pegasus family of models performs better than the other models.

For our system, we, therefore, fine-tune Pegasus on our training set composed of 28k input-output pairs handling different use cases. For fine-tuning the model, we use a single NVIDIA V100 GPU for 30 epochs using a learning rate of $1e-4$ and a batch size of 16. It takes around 16 hours for the model to train. For all our experiments, we use the maximum token sequence length of 512 on both the encoder and decoder.

\section{Human Evaluation for comprehension model metrics description}
\label{sec:comp_human_eval_guidelines}

While conducting the human evaluation for the comprehension models, we ask the animators to make a binary decision on Fluency, Coherence, Naturalness and Coverage. Below we provide the definitions we provided to the annotators.

\paragraph{Fluency}
The generated output should be correct with respect to grammar and word choice, including spelling. It should have no datelines, headers, system-internal formatting, capitalization errors, or ungrammatical sentences (e.g., fragments, missing components) that make the text difficult to read.

\paragraph{Coherence}
The generated output should be well structured and well organized. It should not just be a heap of related information or a collection of sentences but should build from sentence to sentence to a well-organized, naturally flowing, coherent body of information.

\paragraph{Naturalness}
The generated output should use natural phrasing and maintain the appropriate tone and level of formality given its content (e.g., the implied relationship between sender and recipient, the topic, etc.).

\paragraph{Coverage}
The generated output should adequately verbalize the information present in the input. Coverage of all details of the most significant details is desired in the generated output.

\section{Augmentations for training data}
\label{sec:augmentations}

Our system consists of a pipeline of ML models that progressively refines the input at each stage. However, some stages may introduce new errors or fail to fix the errors that they were supposed to fix. The comprehension model is the last stage of the pipeline, and it must address the remaining issues or the new issues introduced by the earlier stages of the pipeline. Therefore, to introduce these capabilities in the comprehension model, we add augmentations to the training dataset of the comprehension model. 

While preparing the training dataset for fine-tuning the \textsc{Comp-FT} model, we generate new training examples by adding augmentations to the input and output of the initial dataset prepared by human annotators. Table \ref{tab:augmentations} shows some of the augmentations we apply. It consists of three columns, showing the augmentation type, the issue it addresses, and its definition. We have four categories in the types of issues we address:

\indent \textbf{ASR issues}: These are issues that were caused by the ASR system, such as incorrectly transcribing a word with its homophone, i.e., similar sounding word.\\
\indent \textbf{Normalization issues}: These are issues that were caused due to issues in the normalization stages, such as missing inserting the correct punctuations or not removing the filler words. \\
\indent \textbf{User input issues}: These are issues that were present in the user speech and were not handled by the earlier models in the pipeline, such as repetition of information or incomplete information in the input. \\
\indent \textbf{Sensitivity issues}: These are issues that we found during our sensitivity reviews, such as the model behaving differently if a non-western name is present in the input.\\

\begin{table*}[]
    \centering
    \small
    \begin{tabular}{l|c|c|c|c|c|c|c} 
     \toprule
     \textbf{Model} & \textbf{Size} & \textbf{BLEU}	& \textbf{ROUGE-1} & \textbf{ROUGE-2}	& \textbf{ROUGE-L} & \textbf{METEOR} & \textbf{BLEURT} \\
     \midrule
     ProphetNet \cite{qi-etal-2020-prophetnet} & 240M & 40.62 & 70.21 & 48.51 & 62.78 & 57.92 & 0.04\\
     GPT2 \cite{radford2019language} & 345M & 46.68 & 73.23 & 56.14 & 68.41 & 64.45 &  0.13\\
     Megatron-GPT2 \cite{shoeybi2019megatron} & 345M & 43.71 & 72.07 & 50.8 & 65.26 & 61.77 &  0.12\\
     BART-large \cite{lewis-etal-2020-bart} & 406M & 49.63 & 72.25 & 53.72 & 67.38 & 66.47 & 0.21 \\
     Pegasus-Base \cite{zhang2020pegasus} & 568M & \textbf{55.98} & 73.3 & 57.47 & 73.05 & 68.95 &  0.25\\
     BigBird-Pegasus-A \cite{zaheer2020big} & 576M & 45.92 & 72.6 & 52.33 & 66.37 & 63.1 & \textbf{0.49}\\
     Pegasus-Large \cite{zhang2020pegasus} & 770M & 53.06 & \textbf{79.33} & \textbf{62.93} & \textbf{74.87} & \textbf{69.71} & 0.26\\
     T5-Large \cite{2020t5} & 770M & 54.23 & 78.81 & 60.03 & 72.27 & 68.82 & 0.24 \\
     GPT3-Neo \cite{black2021gpt} & 1.3B & 45.58 & 73.2 & 52.8 & 67.03 & 63.13 & 0.13\\
    \bottomrule
    \end{tabular}
    \caption{Performance comparison of different fine-tuned Comprehension models on automated metrics.}
    \label{tab:comprehension_auto}
\end{table*}

\begin{table*}[]
    \centering
    \small
    \begin{tabular}{c|c|p{0.5\linewidth}} 
     \toprule
     \textbf{Augmentation} & \textbf{Issue} & \textbf{Definition} \\
     \midrule
     Homophones & ASR & Swap random words in the input with their homophones.\\
     \midrule
     Filler words addition & Normalization & Randomly insert filler words such as uh, um, etc. in the input.\\
     \midrule
     Removing periods and commas & Normalization & Randomly joining sentences by removing the period punctuation in the input. This also helps in making the model learn to generate not very-long sentences.\\
     \midrule
     Content repetition & Noisy User Input & Repeat random words, phrases and sentences in the input so that the model learns to remove repeated information.\\
     \midrule
     Random word removal/addition &  Noisy User Input & Randomly add or remove certain words/phrases (non-entity) in the input so that the model can learn to deal with such noises.\\
     \midrule
     Sentence shuffle &  Noisy User Input & Change the order of certain sentences in the input so that the model can learn to deal with incoherent input.\\
     \midrule
     Gender-neutral rewrite & Sensitivity & Rewrite both the inputs and outputs to a gender-neutral version so that the model does not behave differently for such cases.\\
     \midrule
     Name and date change & Sensitivity & Randomly modify names to non-western names in both input and output so that the model does not behave differently for such cases.\\
     
     \bottomrule
    \end{tabular}
    \caption{A subset of the augmentations used to add more examples to the training data.}
    \label{tab:augmentations}
\end{table*}

\end{document}